\documentclass[letterpaper]{article} 
\usepackage{aaai24}
\usepackage{times}  
\usepackage{helvet}  
\usepackage{courier}  
\usepackage[hyphens]{url}  
\usepackage{graphicx} 
\usepackage{subcaption}
\usepackage{natbib}  
\usepackage{caption} 
\usepackage{algorithm}
\usepackage{algorithmic}

\usepackage{newfloat}
\usepackage{listings}
\DeclareCaptionStyle{ruled}{labelfont=normalfont,labelsep=colon,strut=off} 
\floatstyle{ruled}
\newfloat{listing}{tb}{lst}{}
\floatname{listing}{Listing}
\title{Finetune-Informed Pretraining Boosts Downstream Performance}
\author {
    Atik Faysal\textsuperscript{\rm 1},
    Mohammad Rostami\textsuperscript{\rm 1},
    Reihaneh Gh. Roshan\textsuperscript{\rm 2},
    Nikhil Muralidhar\textsuperscript{\rm 2},
    Huaxia Wang\textsuperscript{\rm 1}
}
\affiliations {
    \textsuperscript{\rm 1} Department of Electrical and Computer Engineering, Rowan University, Glassboro, NJ, USA\\
    \textsuperscript{\rm 2}2
Department of Computer Science, Stevens Institute of Technology, Hoboken, NJ, USA\\
     faysal24@rowan.edu, rostami23@rowan.edu, rghasemi@stevens.edu, nmurali1@stevens.edu, wanghu@rowan.edu
}

\usepackage{bibentry}

\begin{document}

\maketitle

\begin{abstract}

Multimodal pretraining is effective for building general-purpose representations, but in many practical deployments, only one modality is heavily used during downstream fine-tuning. Standard pretraining strategies treat all modalities uniformly, which can lead to under-optimized representations for the modality that actually matters. We propose Finetune-Informed Pretraining (FIP), a model-agnostic method that biases representation learning toward a designated target modality needed at fine-tuning time. FIP combines higher masking difficulty, stronger loss weighting, and increased decoder capacity for the target modality, without modifying the shared encoder or requiring additional supervision. When applied to masked modeling on constellation diagrams for wireless signals, FIP consistently improves downstream fine-tuned performance with no extra data or compute. FIP is simple to implement, architecture-compatible, and broadly applicable across multimodal masked modeling pipelines.
\end{abstract}

\section{Introduction}

Masked autoencoding has become a strong pretraining recipe across vision, language, and signals \citep{he2022masked, devlin2019bert, huang2022masked}. During self-supervised pretraining, models learn to reconstruct missing parts of the input from the visible context, forming general-purpose representations that transfer well to downstream tasks. Conventionally, all modalities are treated equally during pretraining, with the same masking ratio, loss weighting, and model capacity allocated to each. However, downstream deployment often uses one modality most of the time (e.g., constellation diagrams), making uniform pretraining suboptimal. 

So, we raise the question, can we focus on the target modality during pretraining to improve downstream performance? To answer this, in the context of Automatic Modulation Classification (AMC), we propose Finetune-Informed Pretraining (FIP), a strategy that prioritizes the target modality by (1) masking it more aggressively, (2) weighing its reconstruction loss higher, and (3) assigning it a slightly deeper decoder. These changes steer the shared encoder to form features that remain cross-modally useful while being particularly strong for the intended finetuning channel.

\section{Related Work}

\textbf{Masked modeling:} The Masked Autoencoder (MAE) framework, which masks large portions of the input, has been extended to multimodal learning \cite{bachmann20244m}. Models like MultiMAE \cite{bachmann2022multimae}  learn to reconstruct multiple modalities from a shared latent representation, demonstrating strong cross-modal learning. DenoMAE \cite{11005616} adapted this for wireless signals, uniquely introducing noise as an explicit modality to enhance denoising and representation learning. Our work builds directly on DenoMAE, but challenges its assumption of modality symmetry.

\textbf{AMC:} Recent approaches leverage self-supervised learning to reduce labeled data requirements. MCLHN \cite{10562208} uses masked contrastive learning with hard negatives for robust temporal semantics, while GAF-MAE \cite{10261289} transforms signals into Gramian Angular Field images and applies MAE pretraining. These methods, MM-Net \cite{10356142} and MTAMR \cite{10934715}, exemplify a trend toward sophisticated self-supervision and multimodal fusion for robust, data-efficient AMC.

\section{Methodology}

Our approach, FIP, modifies DenoMAE's pretraining objective to prioritize the target modality (constellation diagrams). The DenoMAE backbone includes per-modality patch embeddings, a shared Transformer encoder ($f_{enc}$), and $m$ separate shallow decoders ($f_{dec, m}$). We introduce our modifications, FIP-DenoMAE, below.

\subsection{Asymmetric Masking}

DenoMAE applies a uniform masking ratio ($p_{mask}$) to all modalities, encouraging high-level feature learning. FIP instead uses two distinct ratios: $p_{target}$ for $M_{target}$ and $p_{other}$ for other modalities ($m \neq target$), with $p_{target} > p_{other}$. This creates a more challenging reconstruction task for the target modality, forcing the shared encoder $f_{enc}$ to learn its structure more deeply, while other modalities with lower masking provide richer contextual information for reconstruction.

\subsection{Asymmetric Decoder Architecture}

DenoMAE uses identical, shallow decoders with $L_d$ layers for all modalities. This design is lightweight but forces the encoder $f_{enc}$ to produce representations that are given equal importance across modalities. Following the MAE principle that a lightweight encoder can be paired with a heavier decoder, we apply this asymmetrically. FIP uses a deeper decoder for the target modality, $L_{d,target}$ and retains the shallow decoders for all others, $L_{d,other}$.


\subsection{Weighted Reconstruction Loss}

The DenoMAE multi-modal loss combines the Mean Squared Error (MSE) for each modality
$$ \mathcal{L}_{DenoMAE} = \sum_{m=1}^{n} w_{m}\mathcal{L}_{m} $$
While $w_m$ is included, the baseline implementation treats all $w_m$ as equal.

We set a high weight for the target modality ($w_{target}$) and a lower, non-zero weight for the contextual modalities, $w_{other}$.
$$ \mathcal{L}_{FAP} = w_{target}\mathcal{L}_{target} + \sum_{m \neq target} w_{other}\mathcal{L}_{m} $$

This objective function explicitly guides the shared encoder to prioritize the reconstruction quality of $M_{target}$ over all others.

\begin{figure}[htbp]
    \centering
    \includegraphics[width=0.75\linewidth]{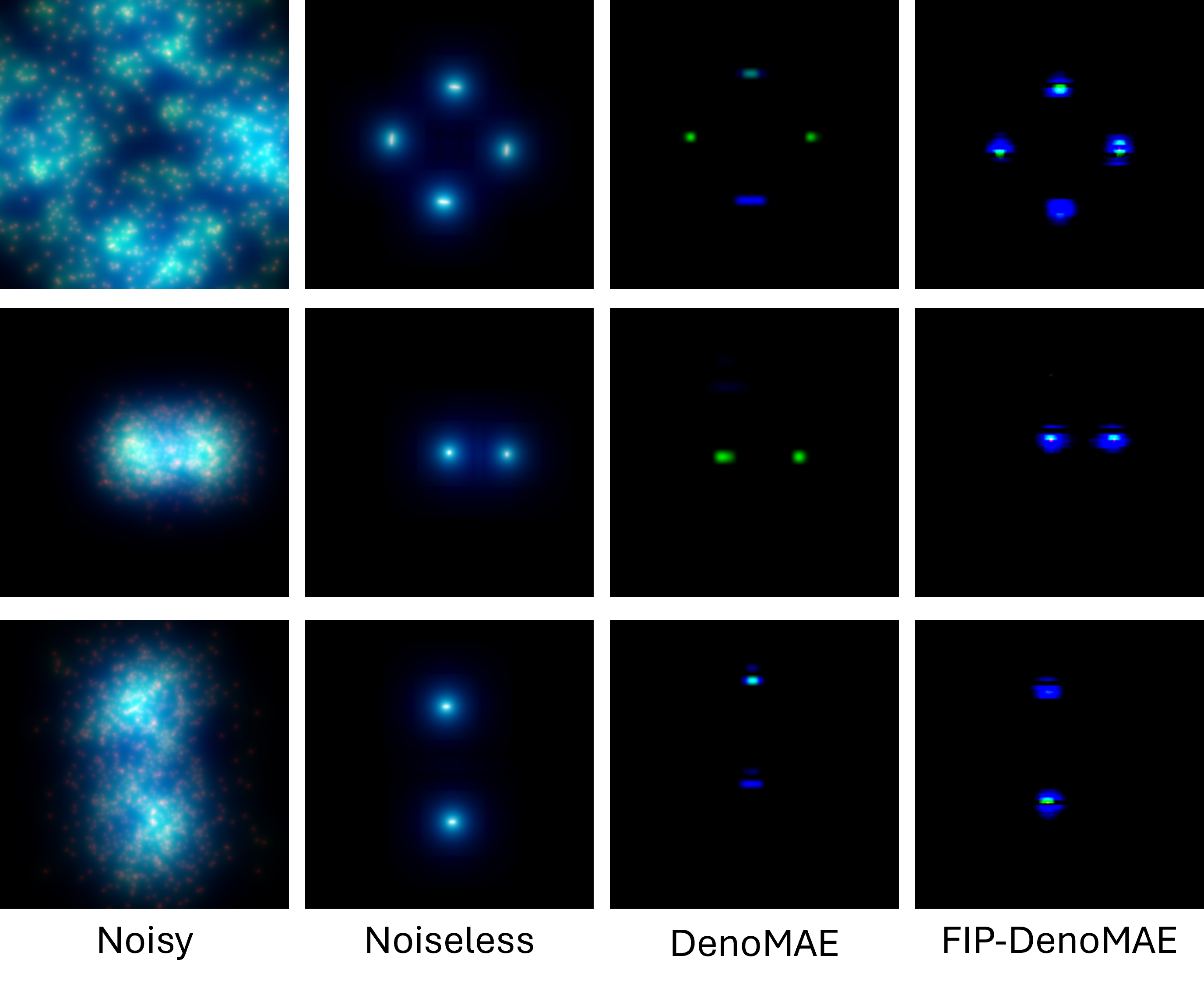}
    \caption{Reconstruction performance of FIP-DenoMAE.}
    \label{fig:recon}
\end{figure}

\section{Experimental Setup}



We follow DenoMAE setup for fair comparison. The dataset contains 10,000 unlabeled samples for pretraining and 1,000 labeled constellation diagrams for 10-class AMC fine-tuning. We use 4 modalities (constellation diagrams, scalograms, raw signals, noise) preprocessed to $3 \times 224 \times 224$ images. The ViT backbone \cite{dosovitskiy2020image} has a 12-layer encoder, three 4-layer decoders ($L_e=12, L_d=4$), embedding dimension $d_{model}=768$, and $16 \times 16$ patches, trained for 100 epochs.

FIP modifies only pretraining to prioritize constellation diagrams. We apply asymmetric masking ($p_{target}=0.80$, $p_{other}=0.60$), deepen the target decoder to $L_{d,target}=8$ (keeping $L_{d,other}=4$), and weight losses as $w_{target}=1.0$ and $w_{other}=0.5$.

\section{Results and Discussion}

In Figure~\ref{fig:recon}, we visualize FIP-DenoMAE and DenoMAE reconstruction results (noisy inputs, clean references, and reconstructed outputs). FIP-DenoMAE effectively denoises constellation diagrams under severe masking with minor artifacts and well-preserved signal structure, outperforming the DenoMAE and demonstrating that target-modality prioritization learns more robust representations.

\begin{figure}[htbp]
    \centering
    \subcaptionbox{DenoMAE}{
        \includegraphics[width=0.45\linewidth]{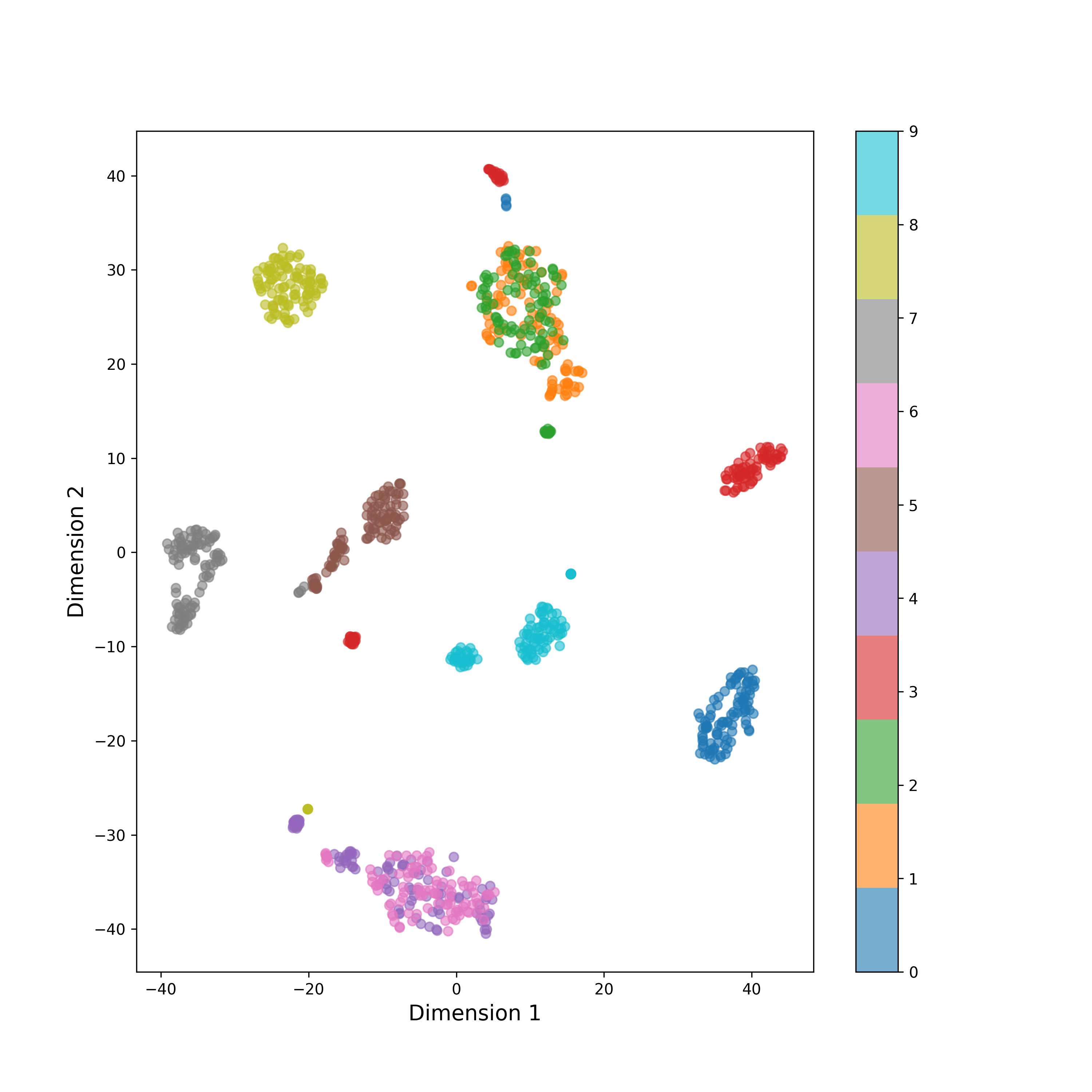}
    }
    \hfill
    \subcaptionbox{FIP-DenoMAE}{
        \includegraphics[width=0.45\linewidth]{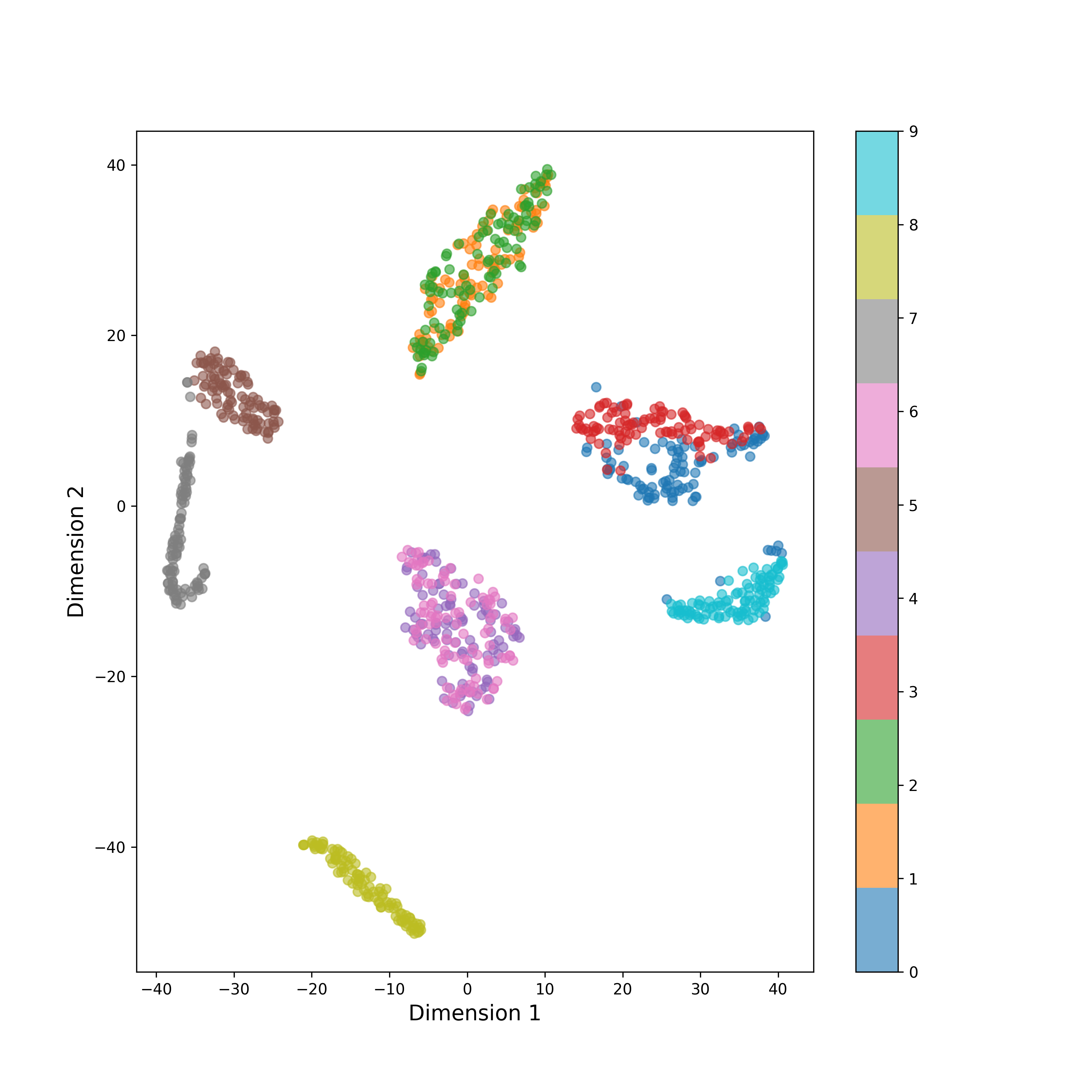}
    }
    \caption{Feature representation using t-SNE.}
    \label{fig:tsne}
\end{figure}

To further analyze representation quality, we visualize the feature space using t-SNE \cite{maaten2008visualizing} in Figure~\ref{fig:tsne}. FIP-DenoMAE shows more distinct class clusters with less overlap compared to DenoMAE, obtaining clearer decision boundaries and better class separability.

\begin{figure}[htbp]
    \centering
    \includegraphics[width=0.85\linewidth]{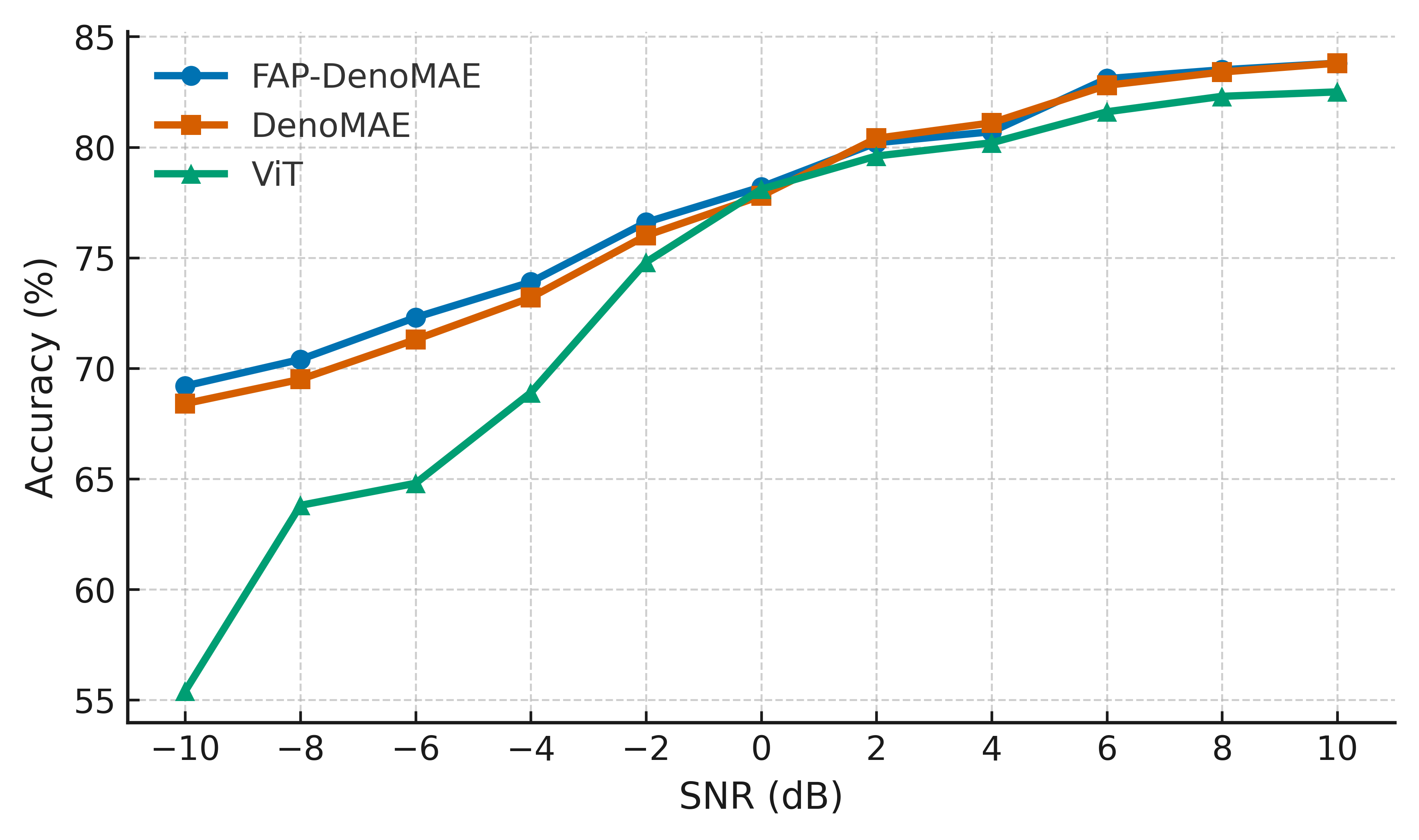}
    \caption{Classification accuracy at different SNRs.}
    \label{fig:snr}
\end{figure}

Figure~\ref{fig:snr} demonstrates that FIP-DenoMAE consistently outperforms both DenoMAE and ViT baselines, with negligible differences at high SNRs (10–6 dB, ~81–84\%). The gap widens significantly in low-SNR regimes, where at -10 dB, FIP-DenoMAE achieves 69.2\% versus DenoMAE's 68.4\% and ViT's 55.4\%, validating that target-modality prioritization substantially improves noise robustness.

\section{Conclusion}

FIP prioritizes the target modality during pretraining through asymmetric masking, decoder depth, and loss weighting. Applied to constellation diagram classification, FIP-DenoMAE consistently outperforms baselines with substantial gains in low-SNR scenarios, without additional data or supervision.

\bibliography{aaai24}

\end{document}